\documentclass{article}

\PassOptionsToPackage{numbers, compress}{natbib}

\usepackage[final]{neurips_2025}

\usepackage[utf8]{inputenc} 
\usepackage[T1]{fontenc}    
\usepackage{hyperref}       
\usepackage{url}            
\usepackage{booktabs}       
\usepackage{amsfonts}       
\usepackage{nicefrac}       
\usepackage{microtype}      
\usepackage{xcolor}         
\usepackage{graphicx}
\usepackage{subfigure}
\usepackage{amsmath,amssymb}

\title{Scaffold Diffusion: Sparse Multi-Category Voxel Structure Generation with Discrete Diffusion}

\author{%
  Justin Jung \\
  Biohub\\
  Redwood City, USA \\
  \texttt{justinsoljung@gmail.com} \\
}

\makeatletter
\renewcommand{\@noticestring}{Accepted at the Third SPIGM Workshop at NeurIPS 2025.}
\makeatother

\begin{document}

\maketitle

\begin{abstract}
Generating realistic sparse multi-category 3D voxel structures is difficult due to the cubic memory scaling of voxel structures and moreover the significant class imbalance caused by sparsity. We introduce Scaffold Diffusion, a generative model designed for sparse multi-category 3D voxel structures. By treating voxels as tokens, Scaffold Diffusion uses a discrete diffusion language model to generate 3D voxel structures. We show that discrete diffusion language models can be extended beyond inherently sequential domains such as text to generate spatially coherent 3D structures.  We evaluate on Minecraft house structures from the 3D-Craft dataset and demonstrate that—unlike prior baselines and an auto-regressive formulation—Scaffold Diffusion produces realistic and coherent structures even when trained on data with over 98\% sparsity. 
We provide an interactive viewer where readers can visualize generated samples and the generation process. Our results highlight discrete diffusion as a promising framework for 3D sparse voxel generative modeling.   
\end{abstract}

\section{Introduction}
Sparse multi-category 3D voxel structures are important data structures in many applications such as computer vision and robotics, entertainment and games, and environment simulation and modeling. However, accurate and realistic generation of sparse multi-category 3D voxel structure come with unique challenges: the cubic nature of voxel structures quickly lead to memory limitations and  the sparsity of the data presents a significant class imbalance which makes accurate generation difficult. While there has been extensive work on generative models for 3D structures more generally and also binary voxel structures, we find work on generation of multi-category \textit{and} sparse voxel structures to be limited. Thus we present Scaffold Diffusion, a discrete diffusion based model for sparse multi-category voxel structure generation. We formulate multi-category voxels as tokens and integrate 3D positional encoding into a masked diffusion language model (MDLM) to yield spatially coherent and realistic generated 3D structures. We also release an interactive demo viewer where users can visualize generated samples and also the generation process: \url{https://scaffold.deepexploration.org/}.  

\begin{figure}[t]
    \centering
    \begin{tabular}{cc}
        \includegraphics[width=0.45\columnwidth]{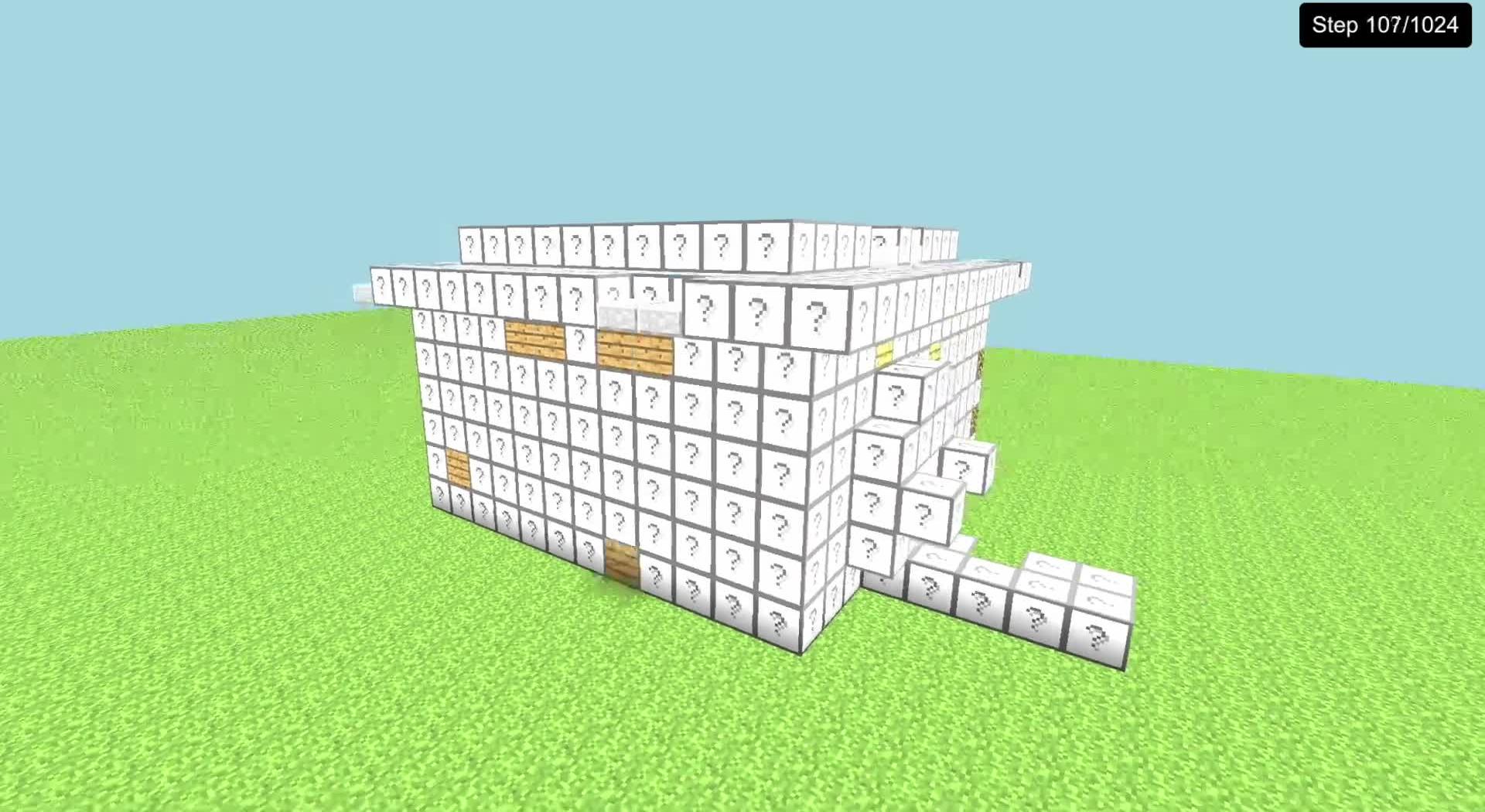} &
        \includegraphics[width=0.45\columnwidth]{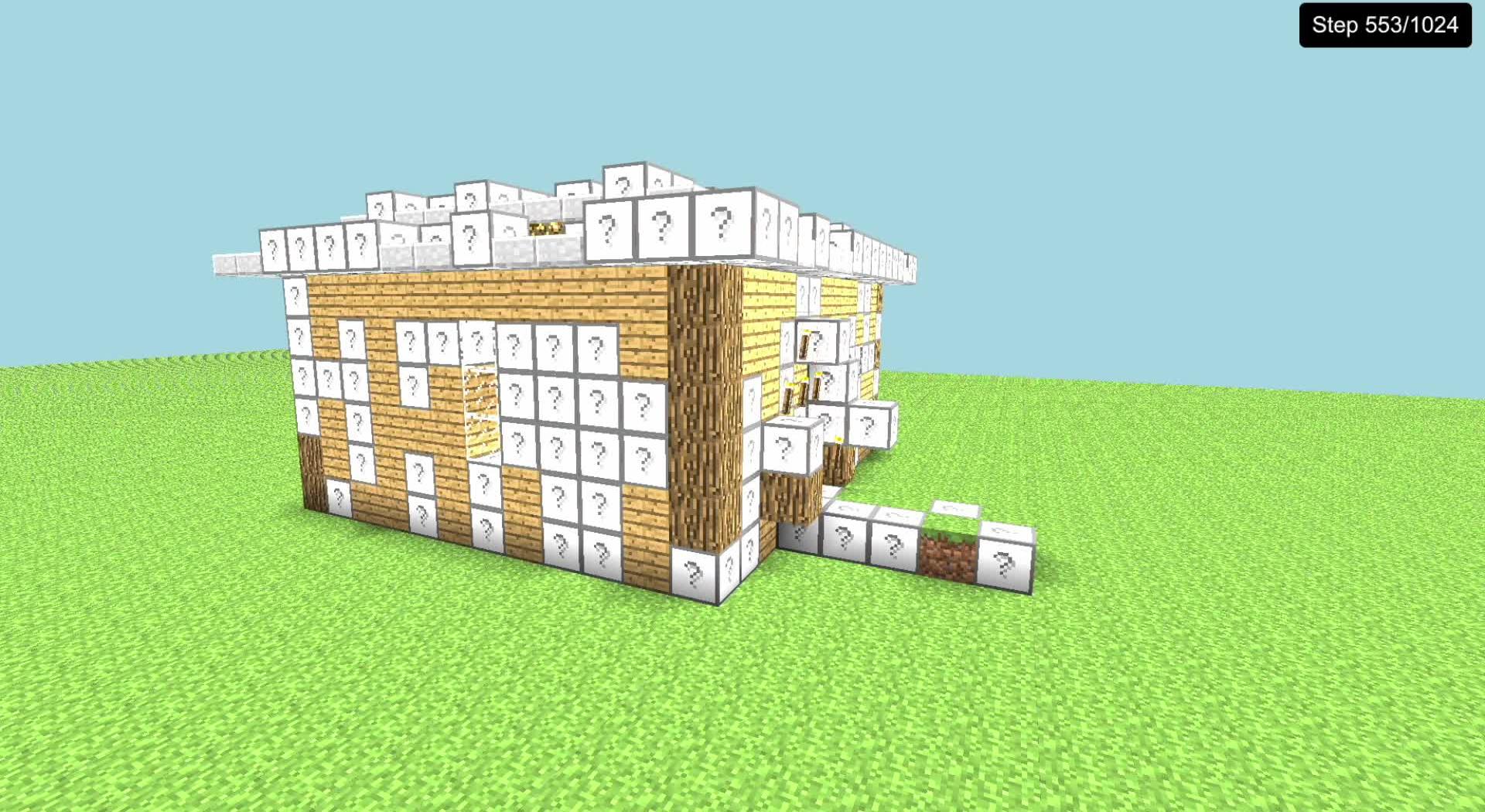} \\
        \small{Early (10\%)} & \small{Mid (50\%)} \\[3pt]
        \includegraphics[width=0.45\columnwidth]{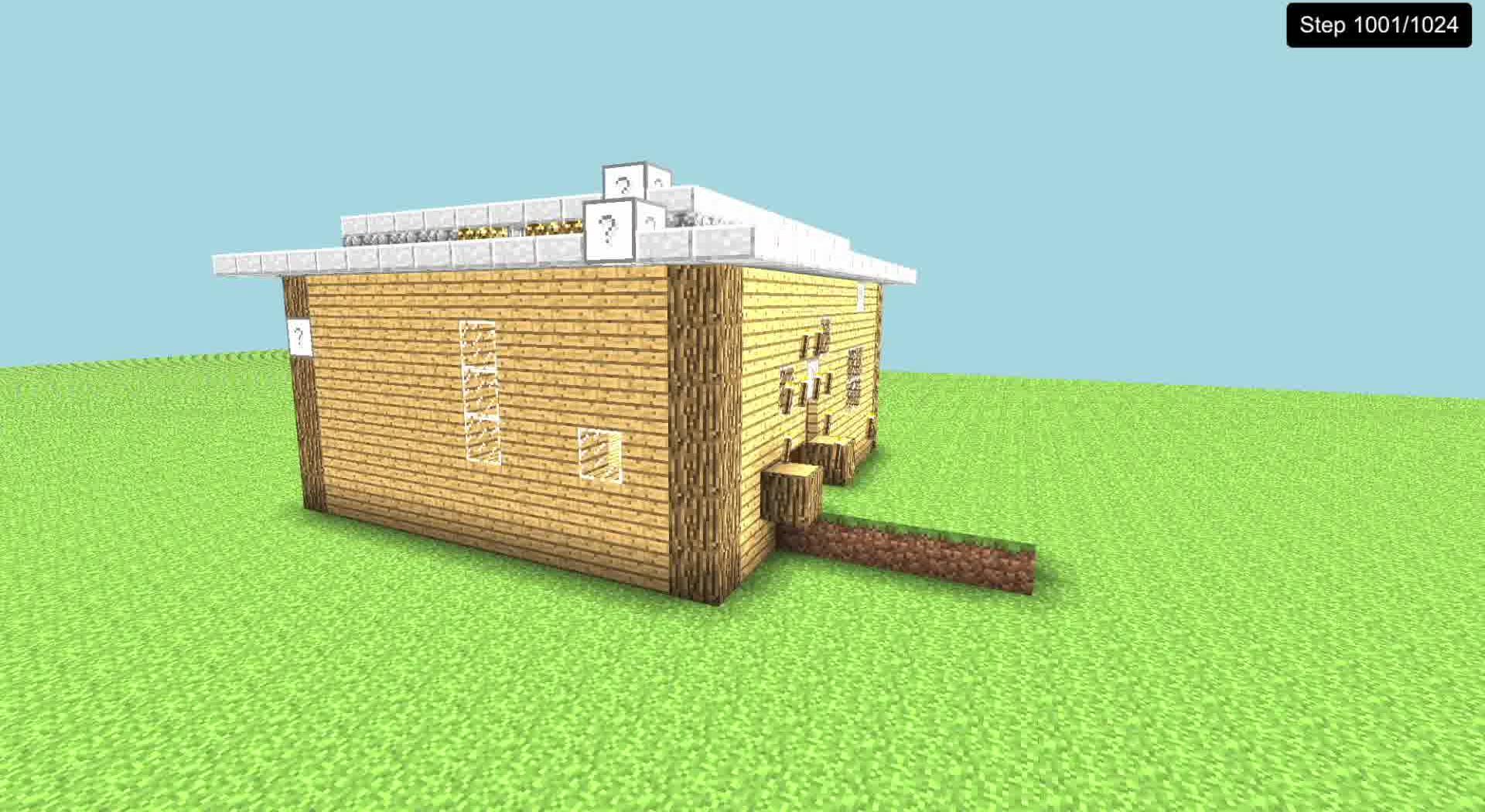} &
        \includegraphics[width=0.45\columnwidth]{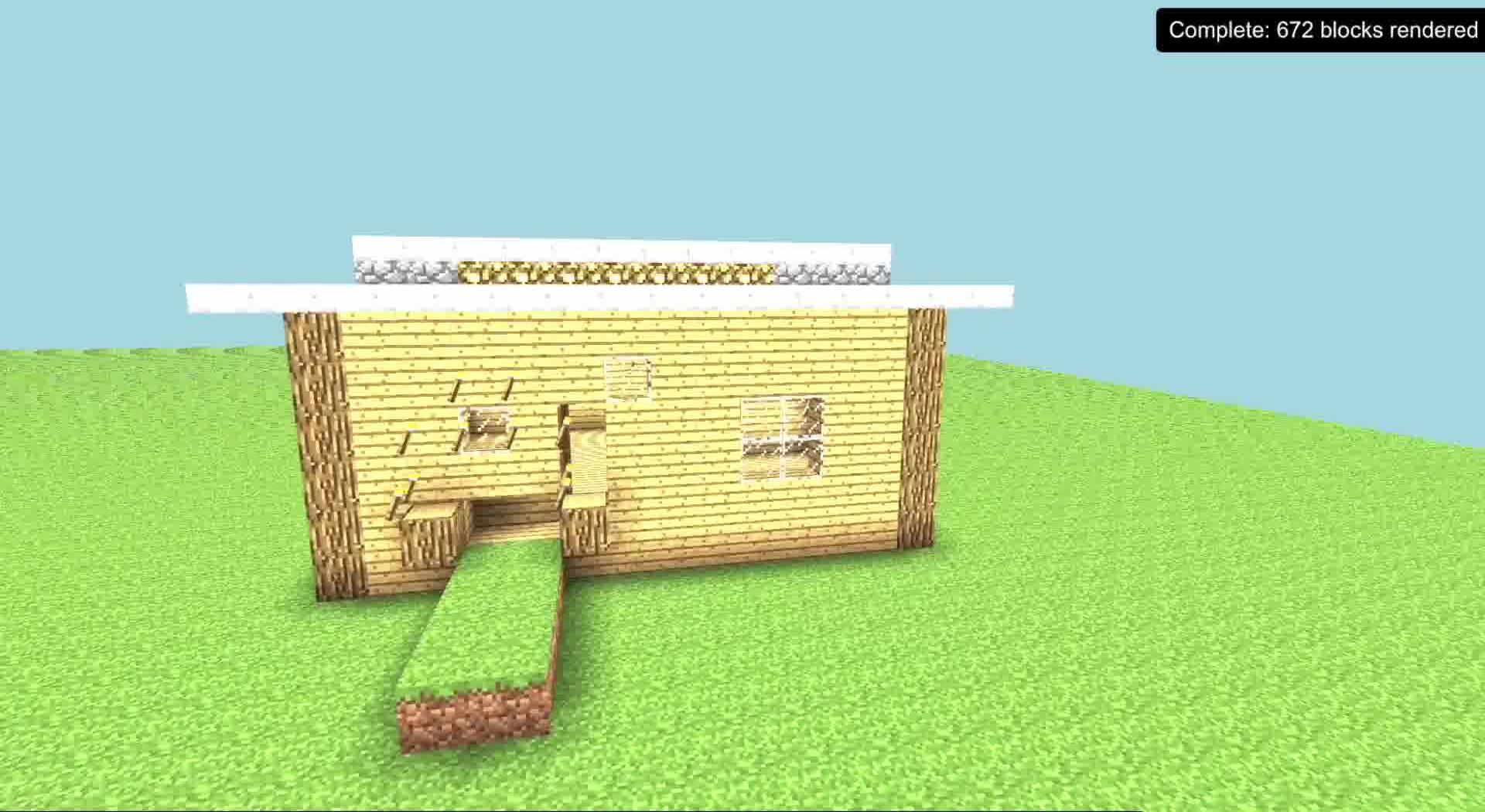} \\
        \small{Late (95\%)} & \small{Complete} \\
    \end{tabular}
    \caption{Progress of voxel structure generation with Scaffold Diffusion.}
    \label{fig:diffusion_progression}
\end{figure}

\section{Related Work}
\subsection{3D and Voxel Generative Models}
Recent advances have shown the promise of 3D generative models. PointVoxelDiffusion~\cite{zhou2021pvd} demonstrates the effectiveness of diffusion models for point cloud generation using point-voxel representations. LiON~\cite{zeng2022lion} introduces a hierarchical latent diffusion model capable of both point cloud and mesh generation. For large scale generation, XCube~\cite{ren2024xcube} presents a hierarchical latent diffusion model for sparse binary-valued voxel maps; for textured structures they apply an off the shelf texture model post-generation. In the domain of floor plan generation, MiDiffusion~\cite{hu2024midiffusion} introduces a mixed continuous-discrete diffusion model that simultaneously predicts categorical labels and geometric features.

Most relevant to our work, Lee et al.~\cite{lee2023diffusion} pioneer the application of discrete diffusion models for 3D categorical data. They apply multinomial diffusion~\cite{hoogeboom2021argmax} to 3D segmentation map generation, and develop both a standalone multinomial diffusion model which simply operates on an entire segmentation map and a two stage VQ-VAE latent multinomial diffusion model which operates on a smaller latent.

\subsection{Minecraft Generative Models}
VoxelCNN~\cite{chen2019voxelcnn} introduces the 3D-Craft dataset of human created minecraft houses and develops an autoregressive 3D convolution network that predicts next block placements given previous block placements. Their model requires previously placed blocks as input and has not been evaluated on complete structure generation, only partial generation.

WorldGAN~\cite{awiszus2021worldgan} attempts large-scale Minecraft world generation using a GAN-based approach with word2vec-like tokenization scheme. However, as the authors acknowledge, their method fails to generate functionally coherent structures such as houses. Alternately, DreamCraft~\cite{earle2024dreamcraft} employs a NeRF-based model and Sudhakaran et al.~\cite{sudhakaran2021growing} uses a neural cellular automata based approach. Both methods however require re-training for each sample, with DreamCraft requiring hours to generate a single sample.

Oasis~\cite{oasis2024} develops a next-frame autoregressive Minecraft world model that generates action-conditioned future frames, allowing the user to interact with the simulated game environment.   

\subsection{Discrete Diffusion Models}
Sohl-Dickstein et al.~\cite{sohldickstein2015deep} first introduces diffusion for discrete spaces with a diffusion process over binary random variables. This work was extended to categorical variables by Multinomial Diffusion~\cite{hoogeboom2021argmax} with a uniform transition process. D3PM~\cite{austin2021d3pm} generalizes Multinomial Diffusion and establishes a discrete diffusion framework with arbitrary transition matrices.

While discrete diffusion has shown promise for generating discrete data, they have generally been considered to have inferior generation quality compared to their autoregressive counterparts. Recent work such as MDLM~\cite{sahoo2024mdlm,shi2025simplifiedgeneralizedmaskeddiffusion} has narrowed the performance gap between discrete diffusion and autoregressive models for text generation through a simplified continuous time ELBO objective, achieving improved sample quality and computational efficiency.

\section{Preliminary}
Diffusion models have been shown to be effective generative models for many data distributions. While diffusion models were initially popularized for continuous data generation, such as image synthesis \cite{ho2020denoising}, discrete diffusion models have been extended to discrete data, such as text \cite{austin2021d3pm}. D3PM \cite{austin2021d3pm} introduces a general framework of discrete diffusion models and evaluates on different forward transition matrices, such as uniform, absorbing state, and discretized gaussian transition matrices. Masked Discrete Language Model (MDLM) is an absorbing state variant of discrete diffusion that has shown effective performance under a simplified training objective \cite{sahoo2024mdlm}. Formally, MDLM operates on discrete token sequences $\mathbf{x} = (x_1, x_2, \ldots, x_L)$ where each token $x_i$ belongs to a finite vocabulary $\mathcal{V}$. The sequence $\mathbf{x} \in \mathbb{Z}^L$ evolves to a sequence of hidden latents $\mathbf{z_t}$ according to the forward corrupting Markov chain defined by the absorbing state transition kernel matrix $Q_t$, where the discrete time forward transition is $q(z_{t} | z_{t-1}) = Cat(z_t ; z_{t-1}Q_t)$. The absorbing state transition kernel matrix $Q_t$ is defined such that each token transitions to the absorbing $[MASK]$ token with probability $\beta_t$ and remains the same with probability $1-\beta_t$. The marginal of the forward process is defined as $q(\mathbf{x_t} | \mathbf{x}) = Cat(\mathbf{x_t} ; \alpha_t \mathbf{x} + (1-\alpha_t) \mathbf{m})$, where $\mathbf{m}$ is the Dirac-delta distribution on the mask token.

\subsection{Training Objective}
Diffusion models aim to maximize a variational lower bound on the log-likelihood of the data distribution; equivalently, they minimize the negative ELBO 
\begin{align}
\mathcal{L} = 
&\; \mathbb{E}_{q} \Bigg[
\underbrace{-\log p_{\theta}(\mathbf{x}|\mathbf{z}_{t(0)})}_{\mathcal{L}_{\text{recons}}} \nonumber \\
&\quad + \underbrace{\sum_{i=1}^{T} D_{\text{KL}}\!\left[ q(\mathbf{z}_{s(i)}|\mathbf{z}_{t(i)}, \mathbf{x}) \,\|\, p_{\theta}(\mathbf{z}_{s(i)}|\mathbf{z}_{t(i)}) \right]}_{\mathcal{L}_{\text{diffusion}}}
\Bigg] \nonumber \\
&\quad + \underbrace{D_{\text{KL}}\!\left[ q(\mathbf{z}_{t(T)}|\mathbf{x}) \,\|\, p_{\theta}(\mathbf{z}_{t(T)}) \right]}_{\mathcal{L}_{\text{prior}}}
\end{align}

MDLM simplifies the variational lower bound and interprets this simplification as a Rao-Blackwellization; the discrete-time formulation is expressed as: 
\begin{align}
\mathcal{L}_{\text{MDLM}} 
&= \sum_{i=1}^{T} \mathbb{E}_{q} \left[
\frac{\alpha_{t(i)} - \alpha_{s(i)}}{1 - \alpha_{t(i)}} 
\log \langle \mathbf{x}_{\theta}(\mathbf{z}_{t(i)}), \mathbf{x} \rangle
\right]
\end{align}
and the continuous-time formulation is expressed as:
\begin{align}
\mathcal{L}_{\text{MDLM}}^{\infty} = \mathbb{E}_q \int_{t=0}^{t=1} \frac{\alpha_t'}{1-\alpha_t} \log\langle x_\theta(z_t, t), x \rangle dt
\end{align}

\begin{figure*}[t]
    \centering
    \setlength{\tabcolsep}{6pt}  
    \renewcommand{\arraystretch}{6}  
    
    \begin{tabular}{ccc}
        \includegraphics[width=0.31\textwidth]{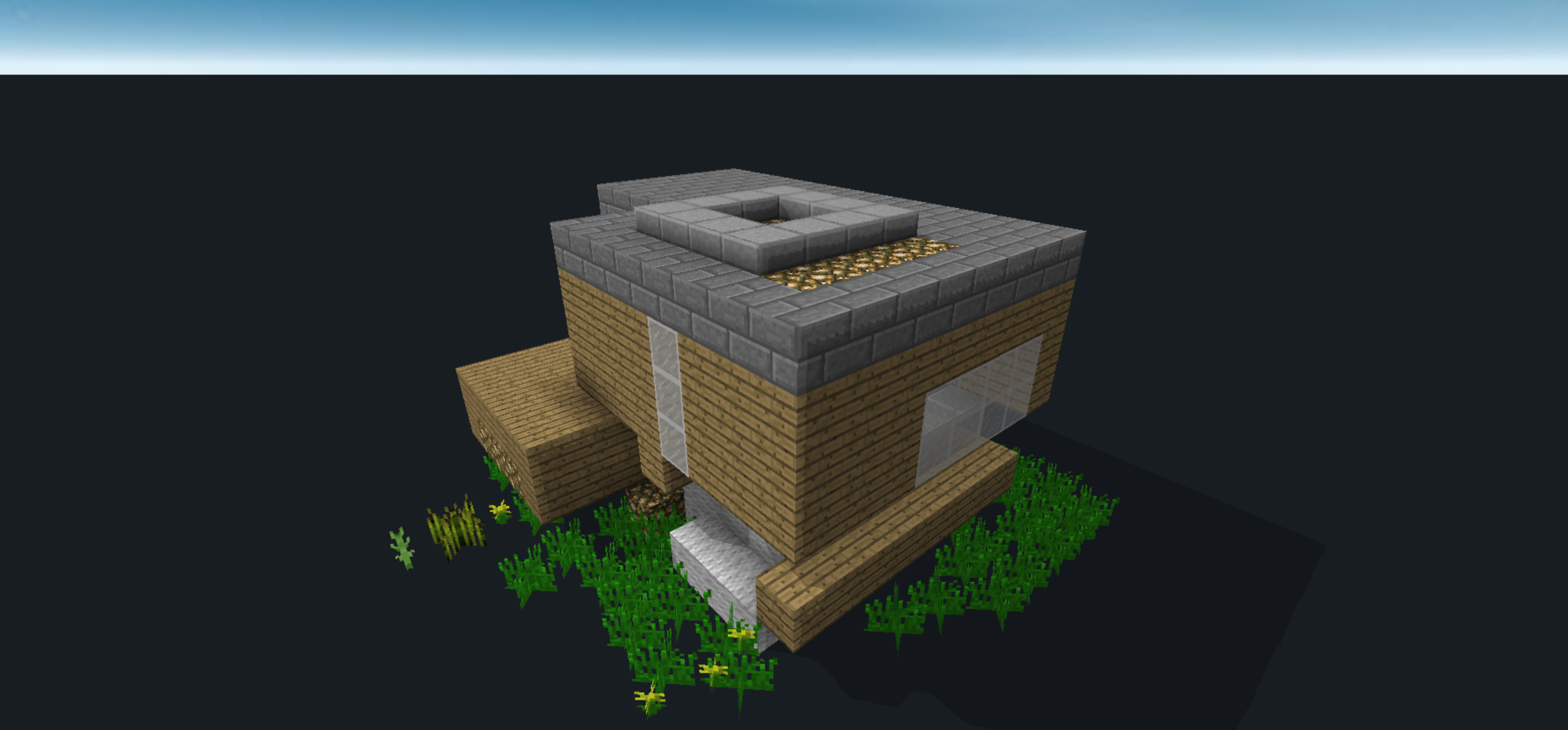} &
        \includegraphics[width=0.31\textwidth]{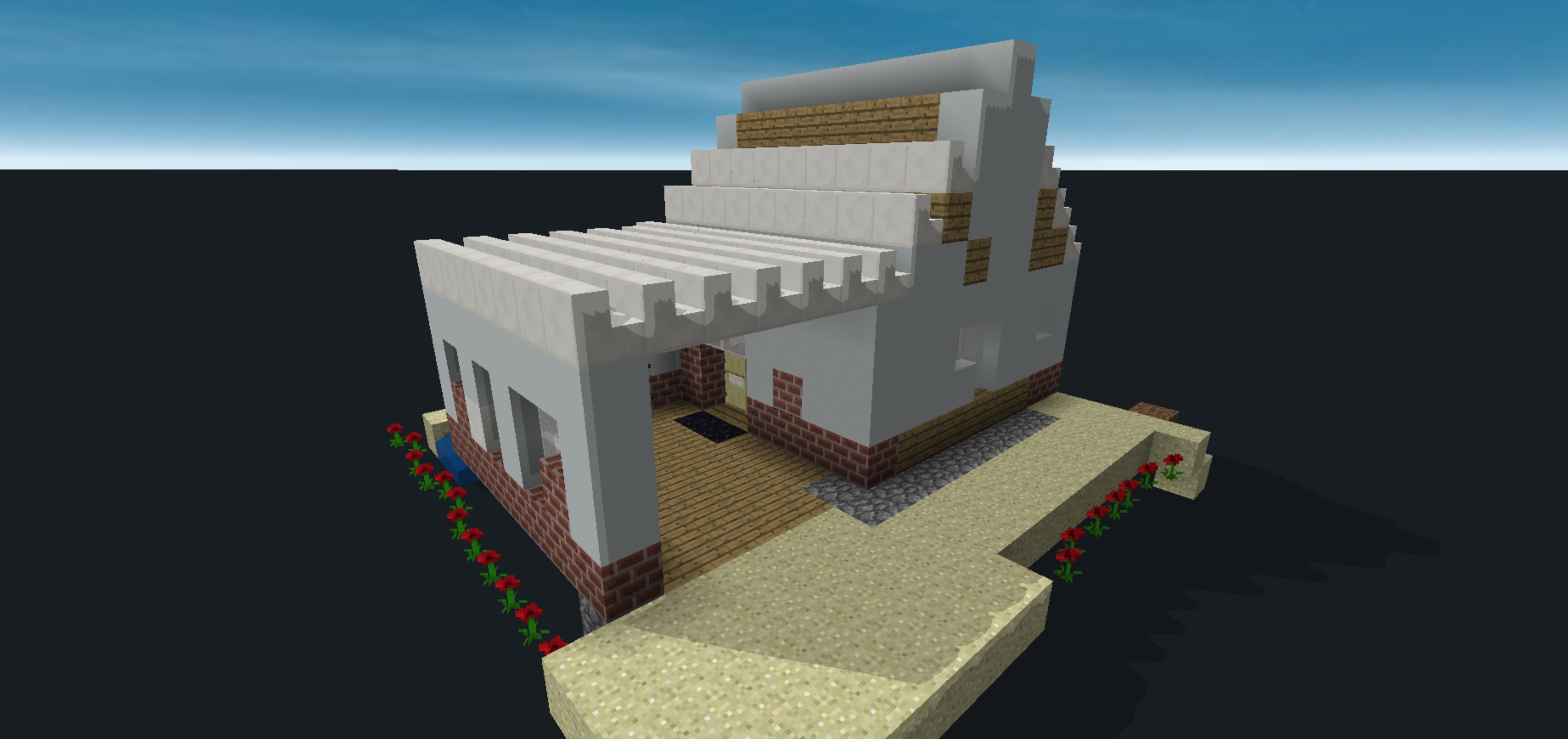} &
        \includegraphics[width=0.31\textwidth]{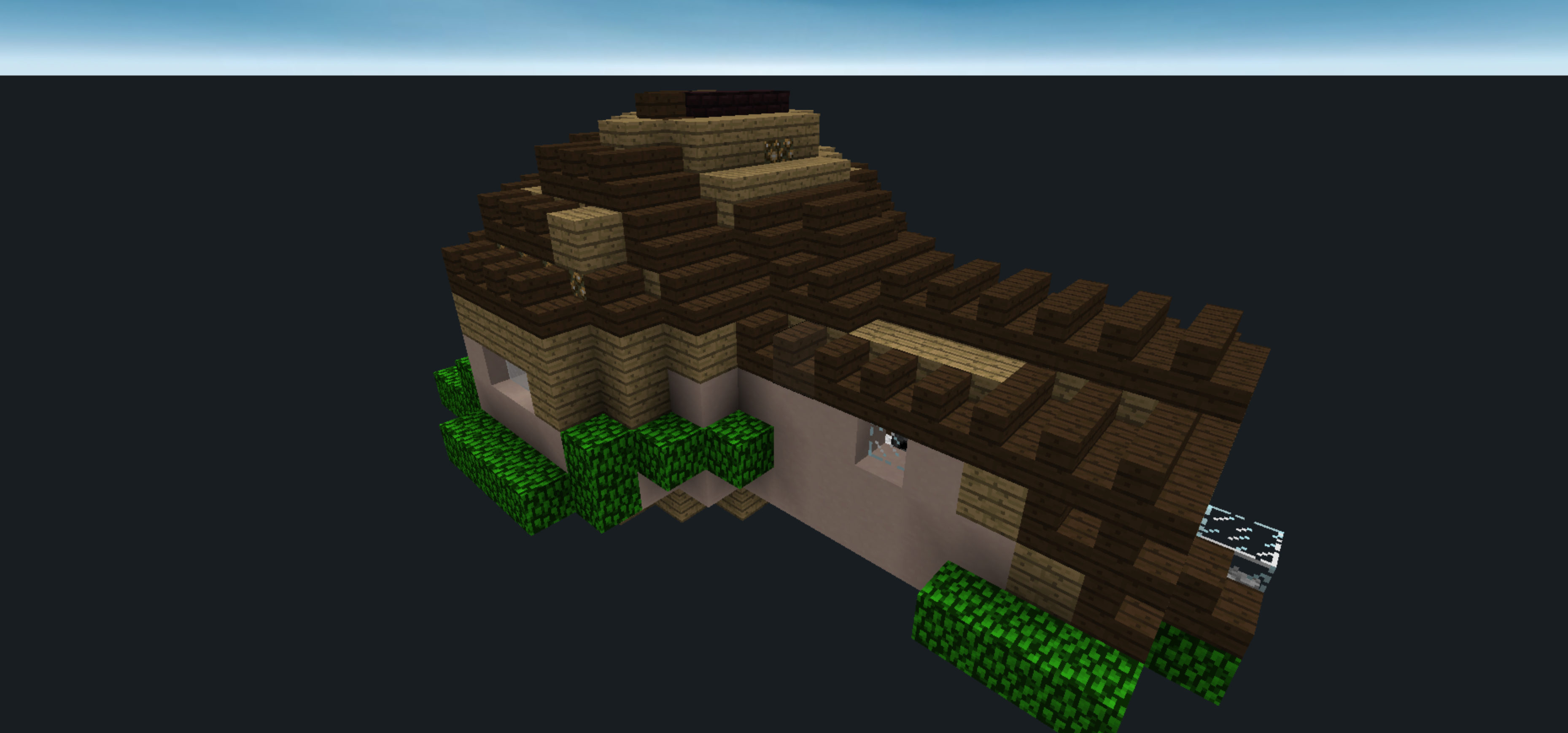} \\
        
        \includegraphics[width=0.31\textwidth]{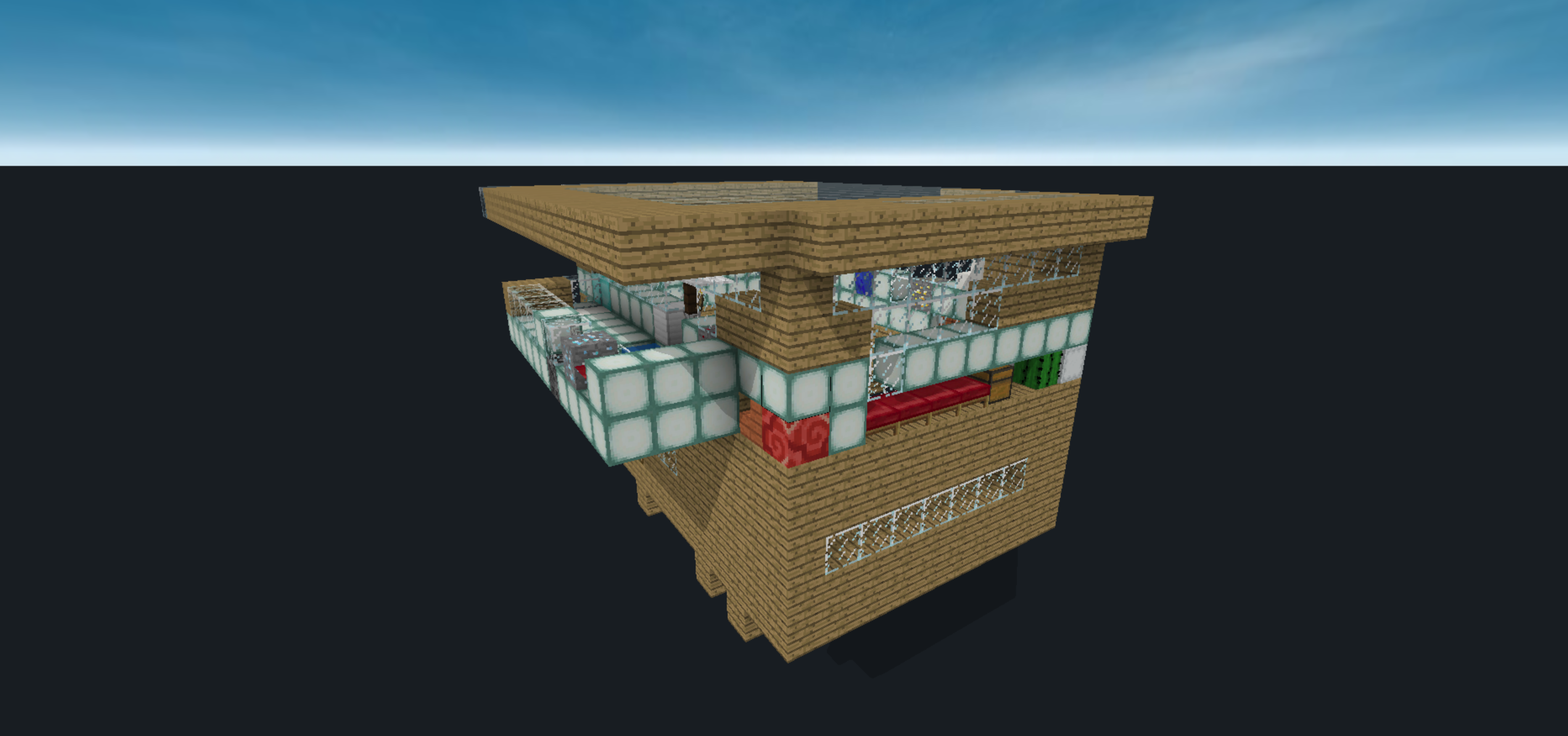} &
        \includegraphics[width=0.31\textwidth]{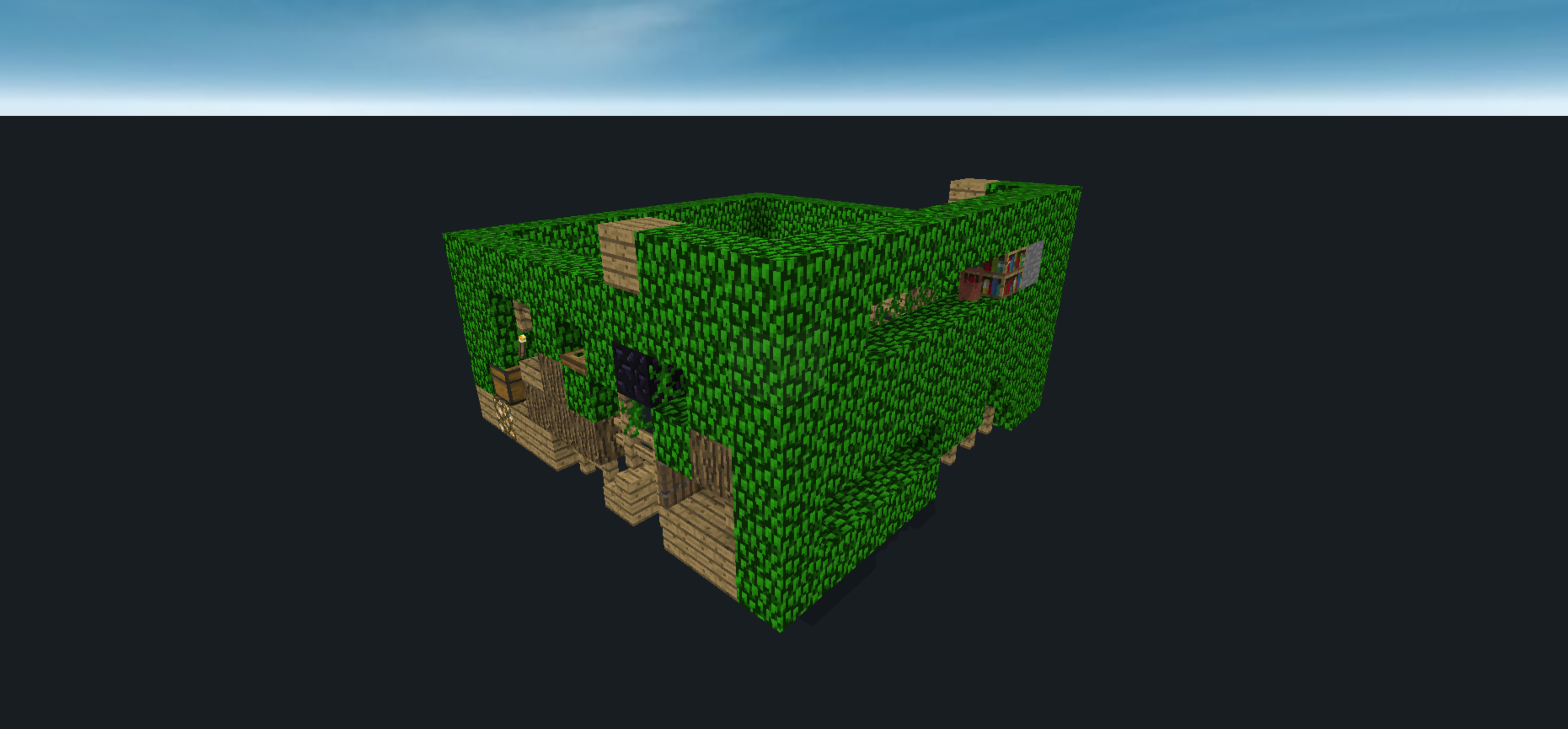} &
        \includegraphics[width=0.31\textwidth]{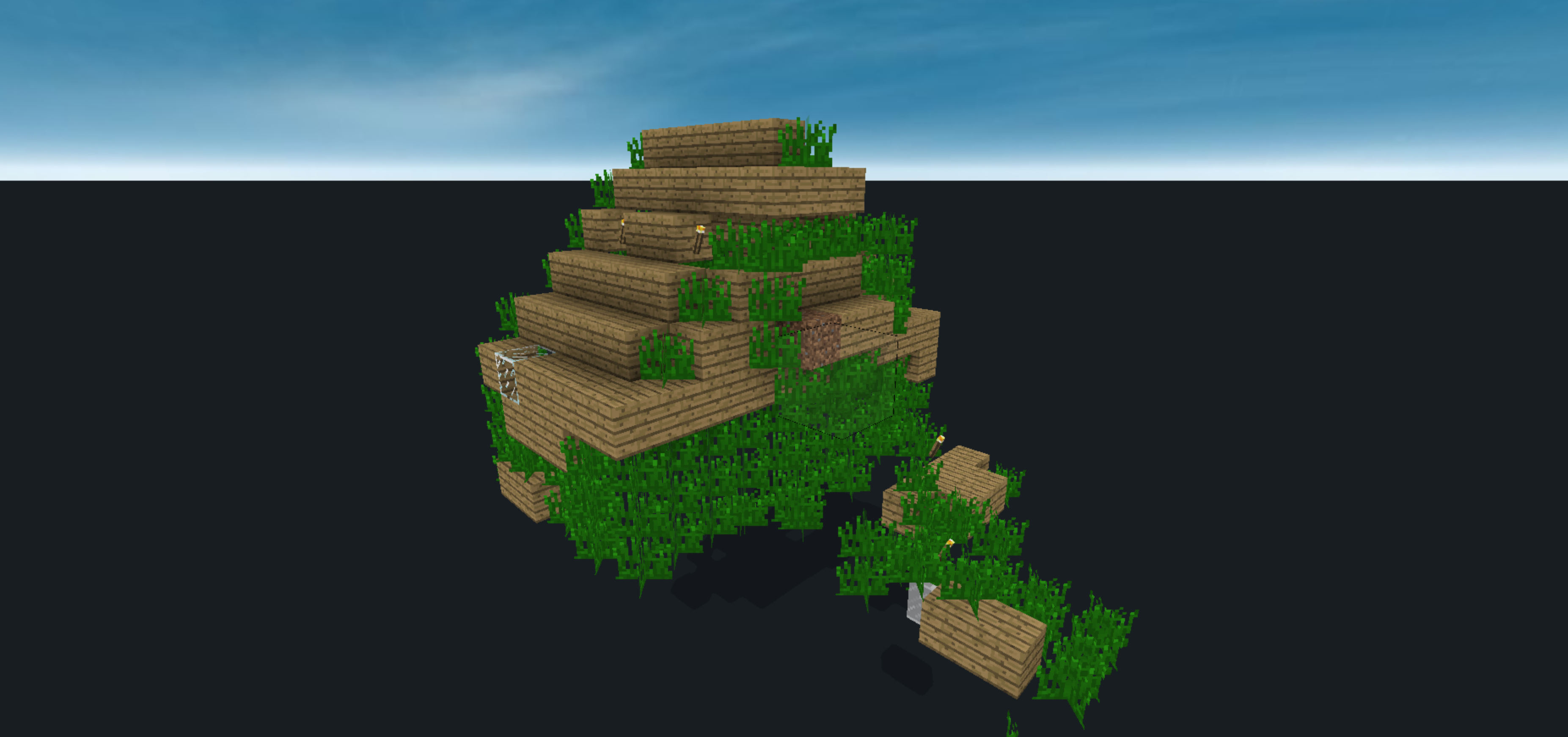} \\
        
        \includegraphics[width=0.31\textwidth]{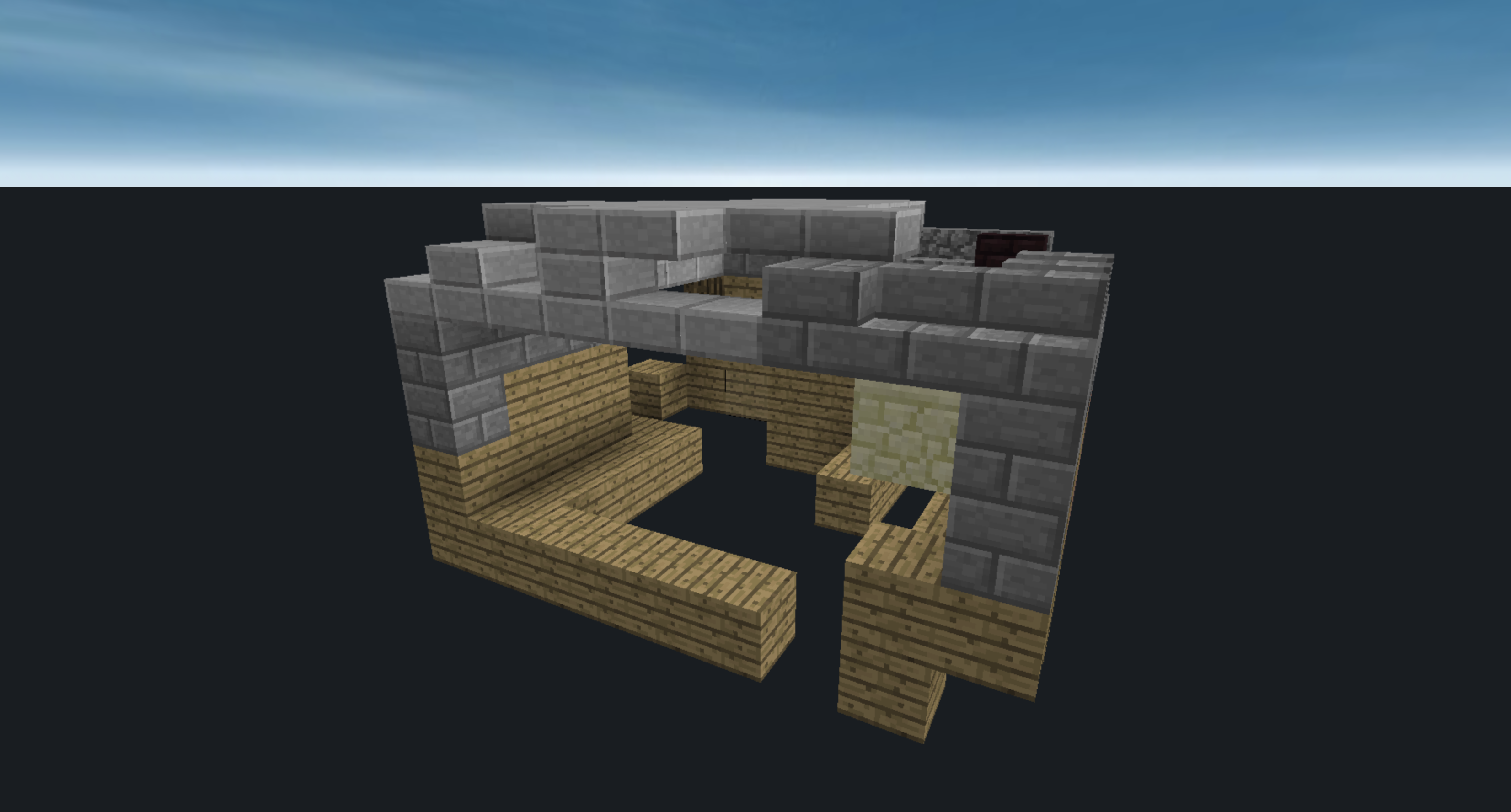} &
        \includegraphics[width=0.31\textwidth]{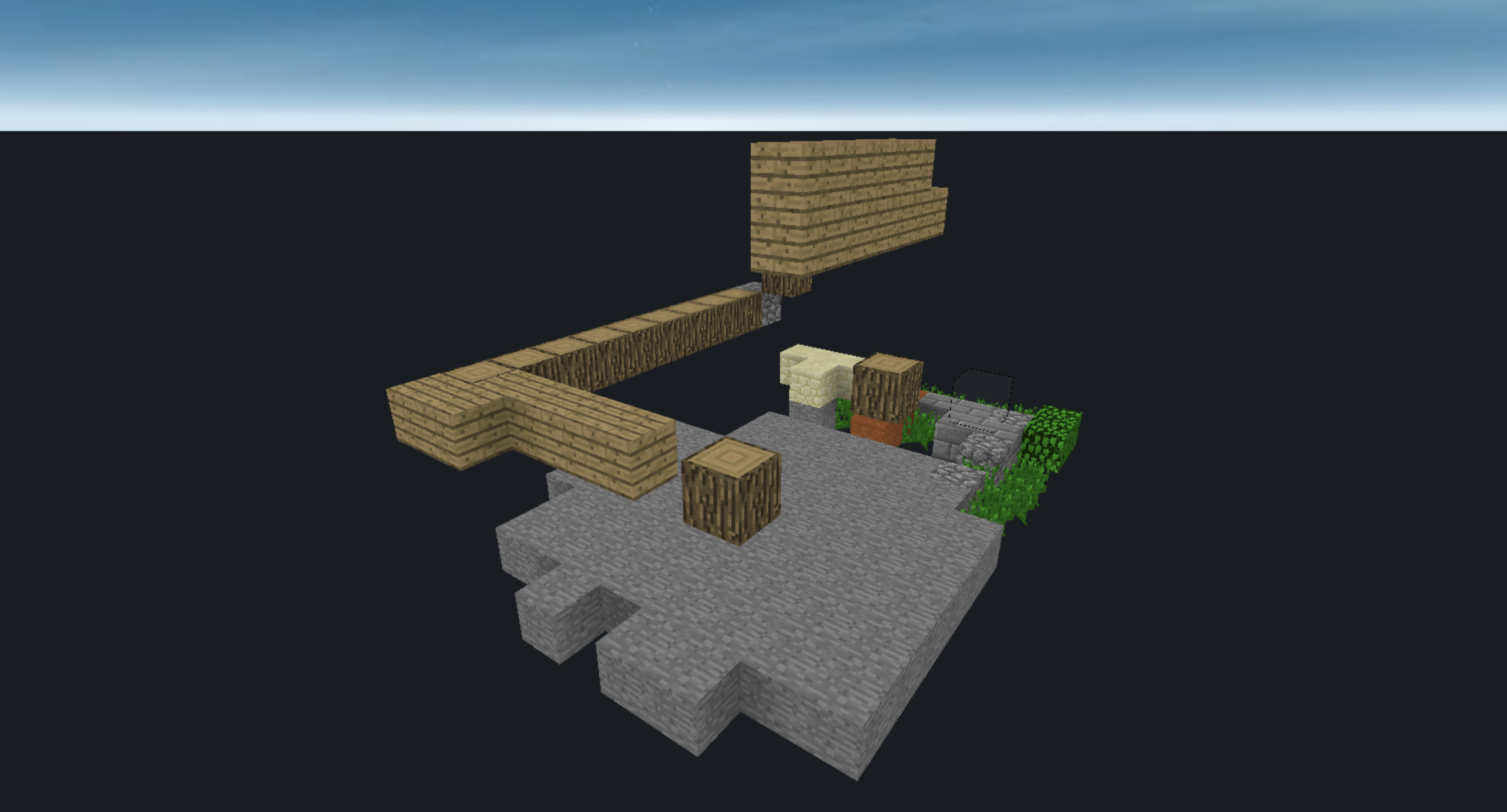} &
        \includegraphics[width=0.31\textwidth]{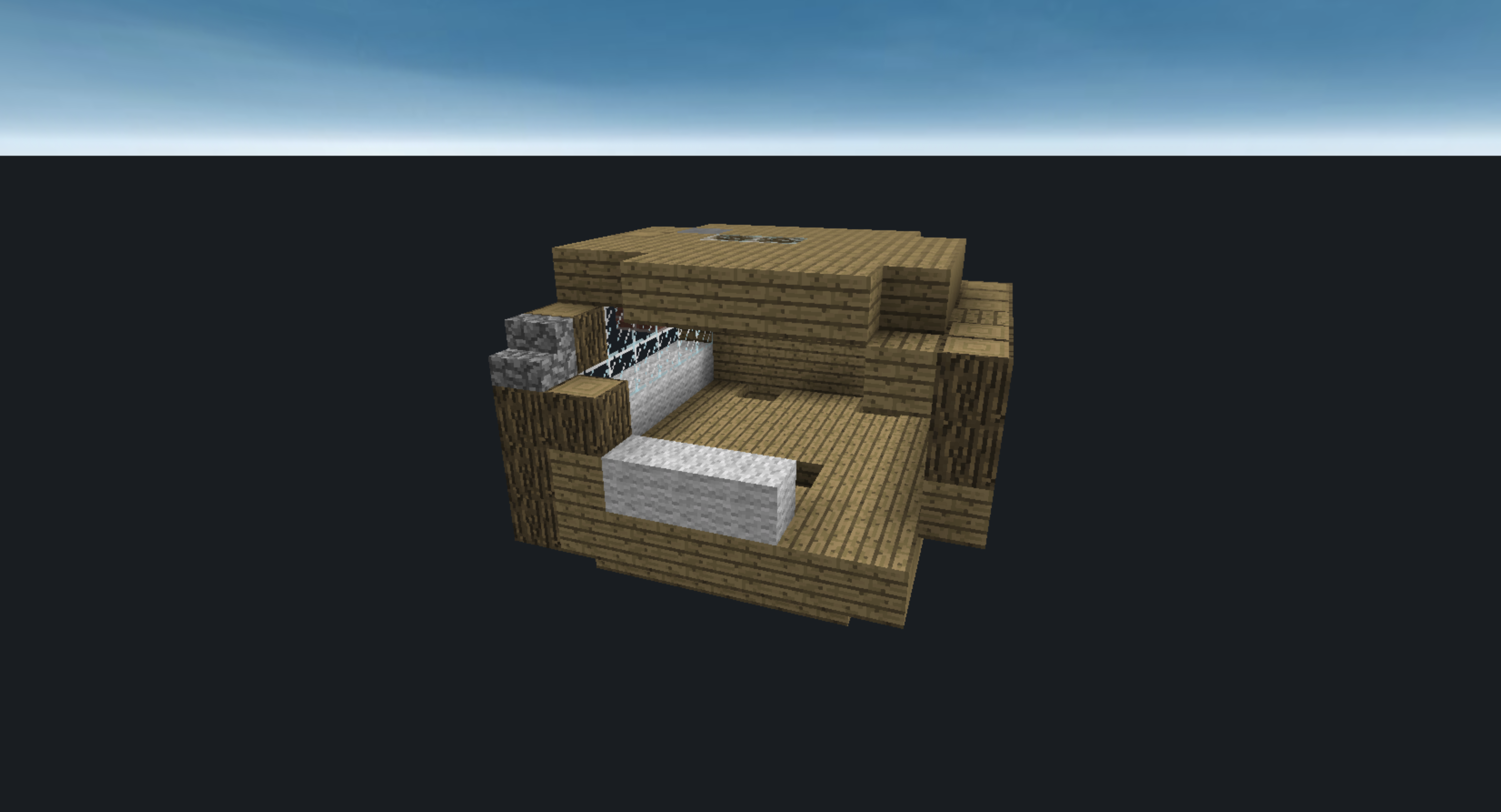} \\
    \end{tabular}
    
    \caption{Sample generation quality comparison between Scaffold Diffusion and baselines; Scaffold Diffusion (top row), autoregressive baseline (middle row), and \cite{lee2023diffusion} (bottom row). While Scaffold Diffusion can generate realistic and functional 3D structures, the autoregressive baseline generates structures dominated by a few block types or structures with implausible block placements. \cite{lee2023diffusion} suffers from an over-representation of background voxels.}
    \label{fig:comparison}
\end{figure*}

\section{Method}
Generating realistic sparse voxel structures poses a unique challenge due to the class imbalance with disproportionate empty background voxels and moreover memory issues due to its cubic ($O(n^3)$) nature. To address these challenges, Scaffold Diffusion conditions on a boolean occupancy map $\mathbf{O} \in \mathbb{Z}^{D\times D\times D}$ and generates a spatially coherent structure for the given occupied voxels. (Learning to first generate the boolean occupancy map is left as potential future work.)

Because of the memory and class imbalance challenges of working with the full voxel map $\mathbf{O} \in \mathbb{Z}^{D\times D\times D}$, we extract only the $k$ many occupied voxels and their corresponding voxel location $\{(x_i,y_i,z_i) \}_{i=1}^{k}$. From these occupied voxels, we extract their locations to define a $L \geq k$ length sequence $\mathbf{x} \in \mathbb{Z}^L$ (potentially appended with a variable number of padding tokens) for the masked diffusion language model to generate. For our model to be spatially aware during its generation process, we integrate 3D positional embedding. We find 3D sinusoidal positional embeddings as in \cite{mo2023dit} effective. Finally, given a generated sequence $\mathbf{x}_0 \in \mathbb{Z}^L$ and the occupied voxel locations, we can re-construct a generated voxel map $\mathbf{X} \in \mathbb{Z}^{D\times D\times D}$ for visualization purposes.      

\section{Experiments}
We evaluate on the 3D-Craft dataset \cite{chen2019voxelcnn}, a dataset of human-created Minecraft houses. The 3D-Craft dataset contains a time-stamped sequence of block placements $\{(x_t,y_t,z_t, id_t) \}_t$. The Minecraft block IDs are integers in the range $[0,255]$ and we have in our dataset $n=253$ possible block ID values, leading to a vocabulary size of $|V| = 253$. Since in many settings a sequence of block placements is not available, we convert the sequences to a voxel structure by placing the block placements in a voxel cube of some pre-defined dimension $\mathbf{X} \in \mathbb{Z}^{D \times D \times D}$.  

We choose the typical sequence length of $L=1024$ and consider voxel cube dimensions of $32^3$ and $64^3$. Our results shown below are samples generated with the voxel cube dimensions $32^3$, but we observe qualitatively similar samples with voxel cube dimensions $64^3$. For our total dataset, we subset structures that contain at most $1024$ occupied blocks and fit within a $32^3$ voxel cube, leading to a total of $1432$ house voxel structures which are on average $98.3$\% background tokens.  

\subsection{Implementation Details}
For our discrete diffusion model, we use the Masked Diffusion Language Model MDLM \cite{sahoo2024mdlm}. We adopt their code and use their model architecture, optimizer, and hyperparameters. Specifically, we use the Diffusion Transformer (DiT) backbone \cite{peebles2023scalable} with $n=12$ blocks and $n=12$ heads and a sequence length of $L=1024$. We use a log-linear noise schedule and for faster inference we use cached updates. We use EMA with $\beta=0.9999$ and AdamW optimizer with a learning rate of $lr= 3\times10^{-4}$, $\beta_1 = 0.9$, $\beta_2 =0.999$ and $\epsilon = 1\times 10^{-8}$ and a constant warm-up of $2500$ steps. We train our models with a maximum of $10^{6}$ steps. We additionally integrate 3D sinusoidal positional embeddings as in \cite{mo2023dit} into our DiT backbone.

All experiments were trained using a single RTX 5090 and each experiment took under 12 hours to finish training. 

\subsection{Baselines}
As a comparable baseline, we implement a VQ-VAE latent multinomial diffusion model as in \cite{lee2023diffusion}. In line with their method, the model is trained with an inverse class-frequency weighted cross-entropy loss to mitigate the effects of class imbalance. We adopt their default hyperparameter settings and their provided code. 

Additionally, to motivate the choice of discrete diffusion as our generative modeling framework, we construct an autoregressive baseline which has the same training setup and backbone model, but with a next-token prediction objective and no timestep conditioning. For this model, we also prepend a $[BOS]$ token to all sequences to allow for sequence generation.

\begin{figure*}[t]
    \centering
    \setlength{\tabcolsep}{6pt}  
    \renewcommand{\arraystretch}{6}  
    
    \begin{tabular}{ccc}
        \includegraphics[width=0.31\textwidth]{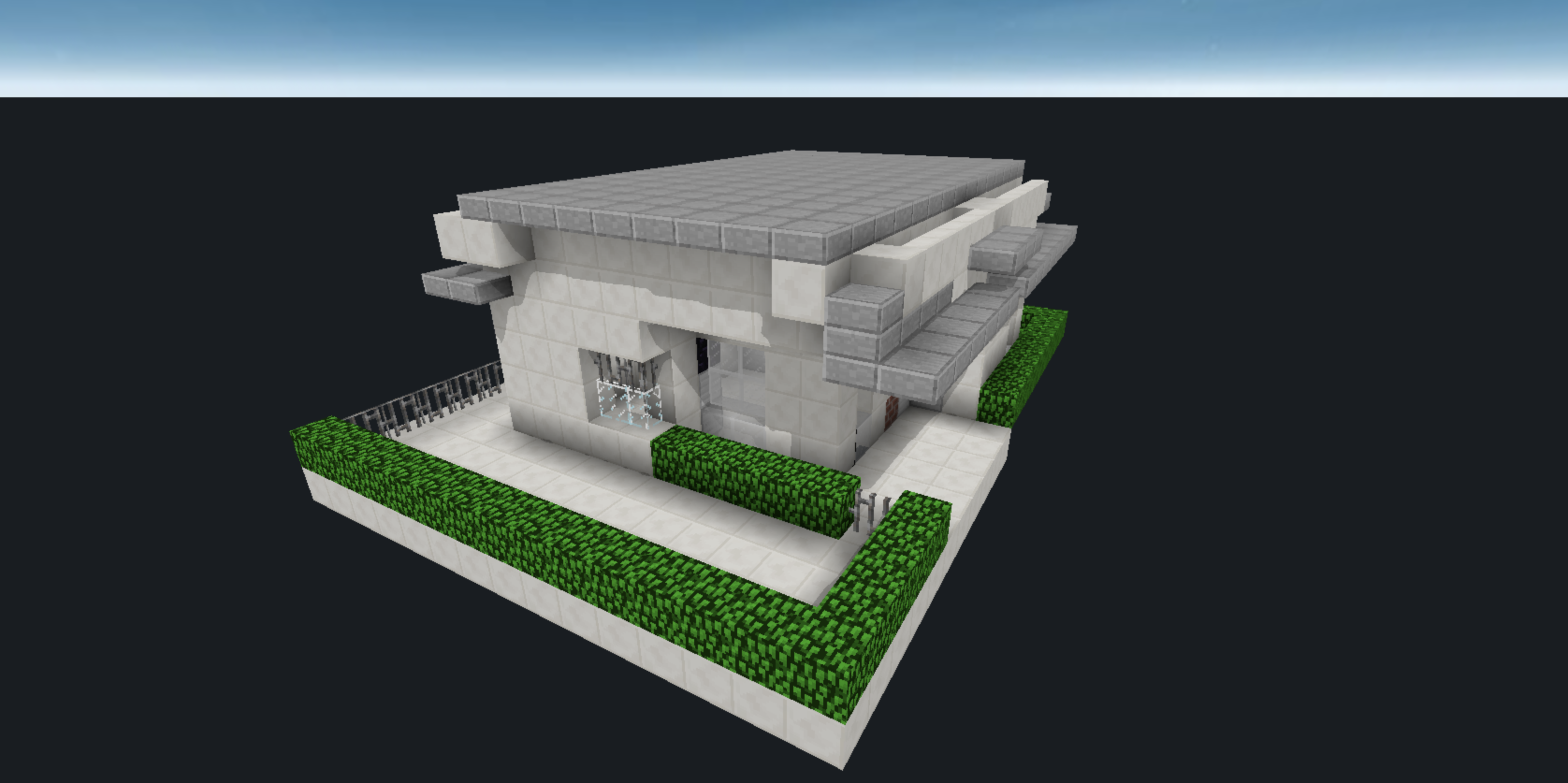} &
        \includegraphics[width=0.31\textwidth]{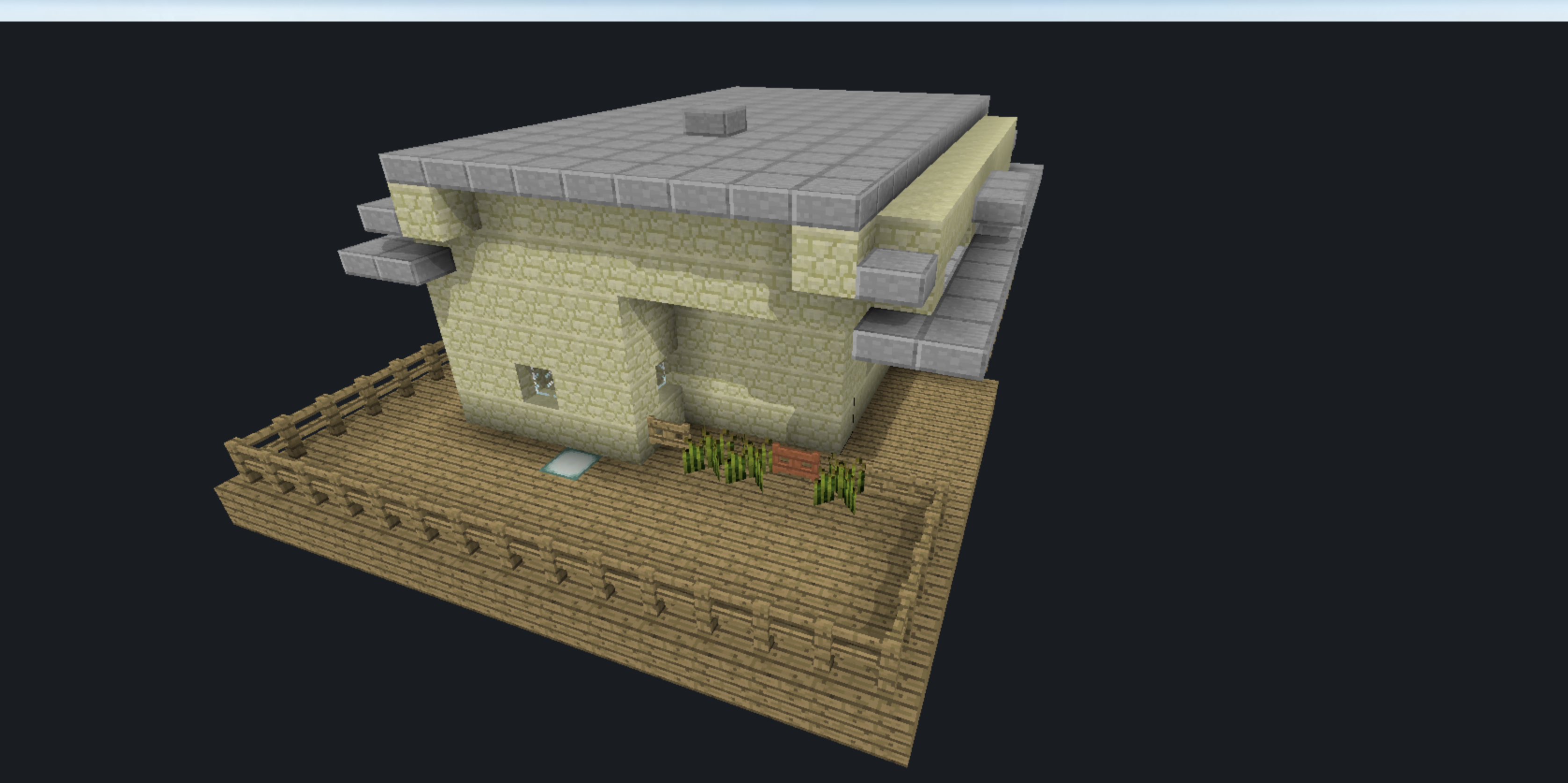} &
        \includegraphics[width=0.31\textwidth]{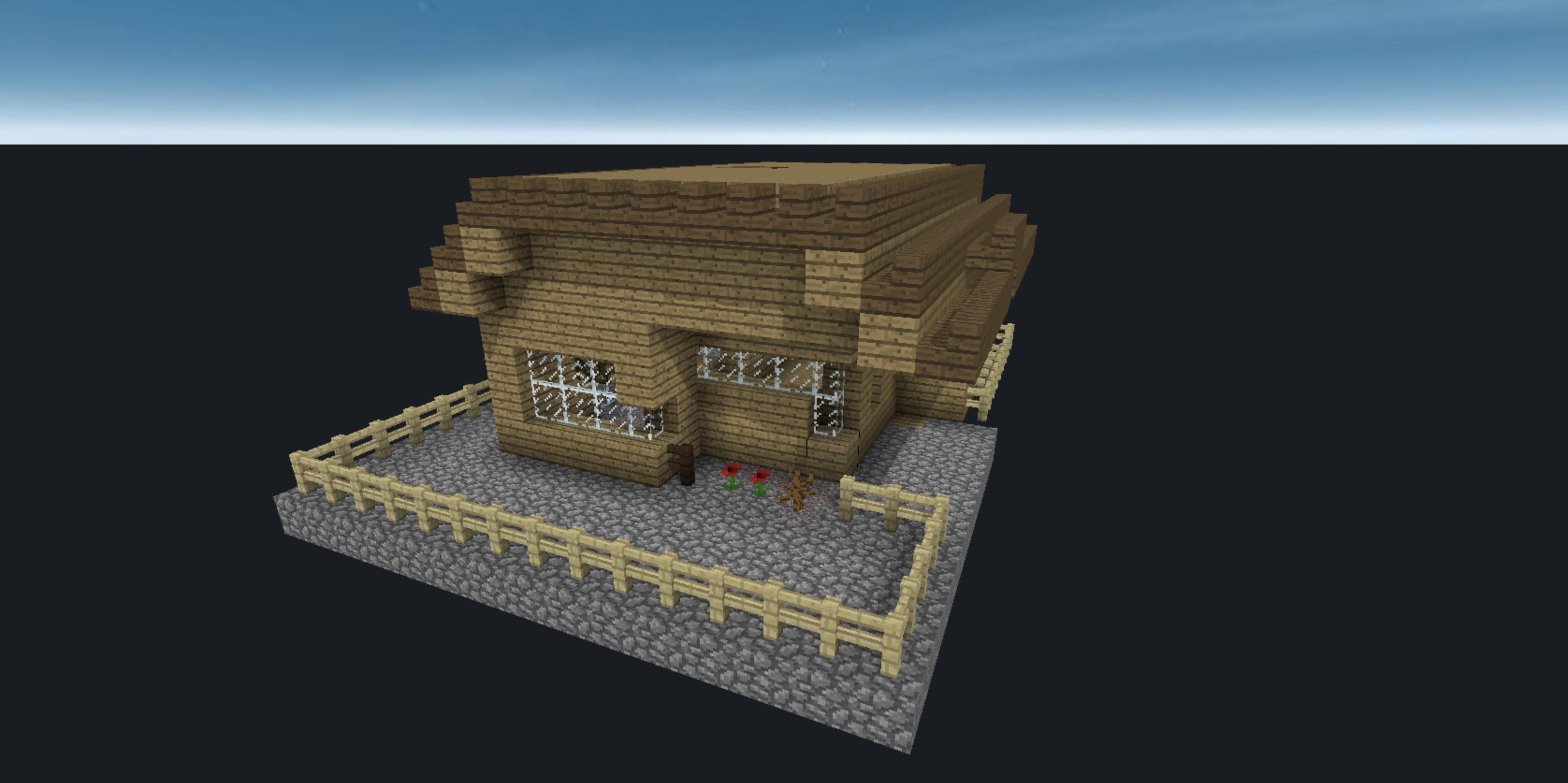} \\[2pt]
        
        \includegraphics[width=0.31\textwidth]{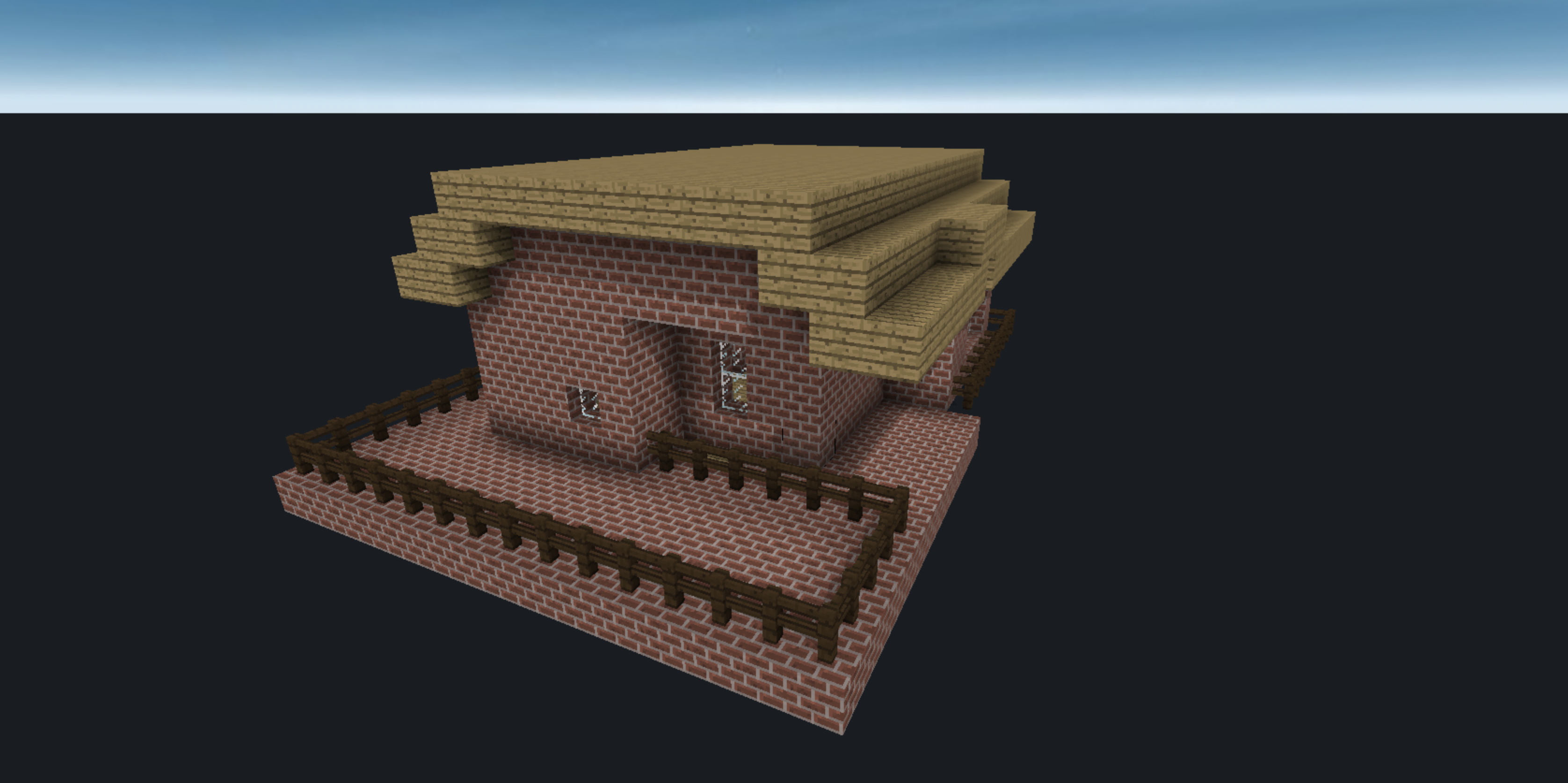} &
        \includegraphics[width=0.31\textwidth]{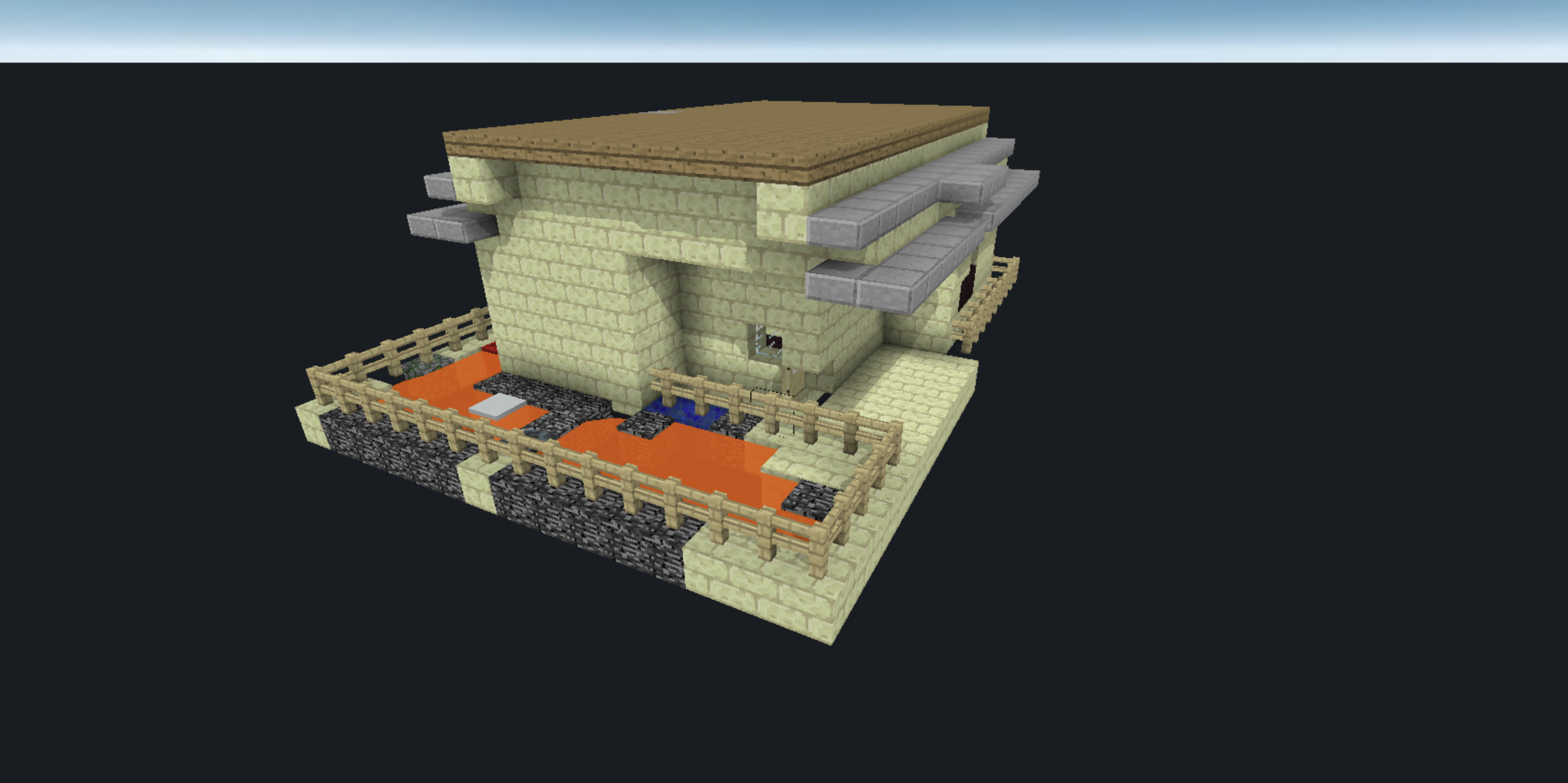} &
        \includegraphics[width=0.31\textwidth]{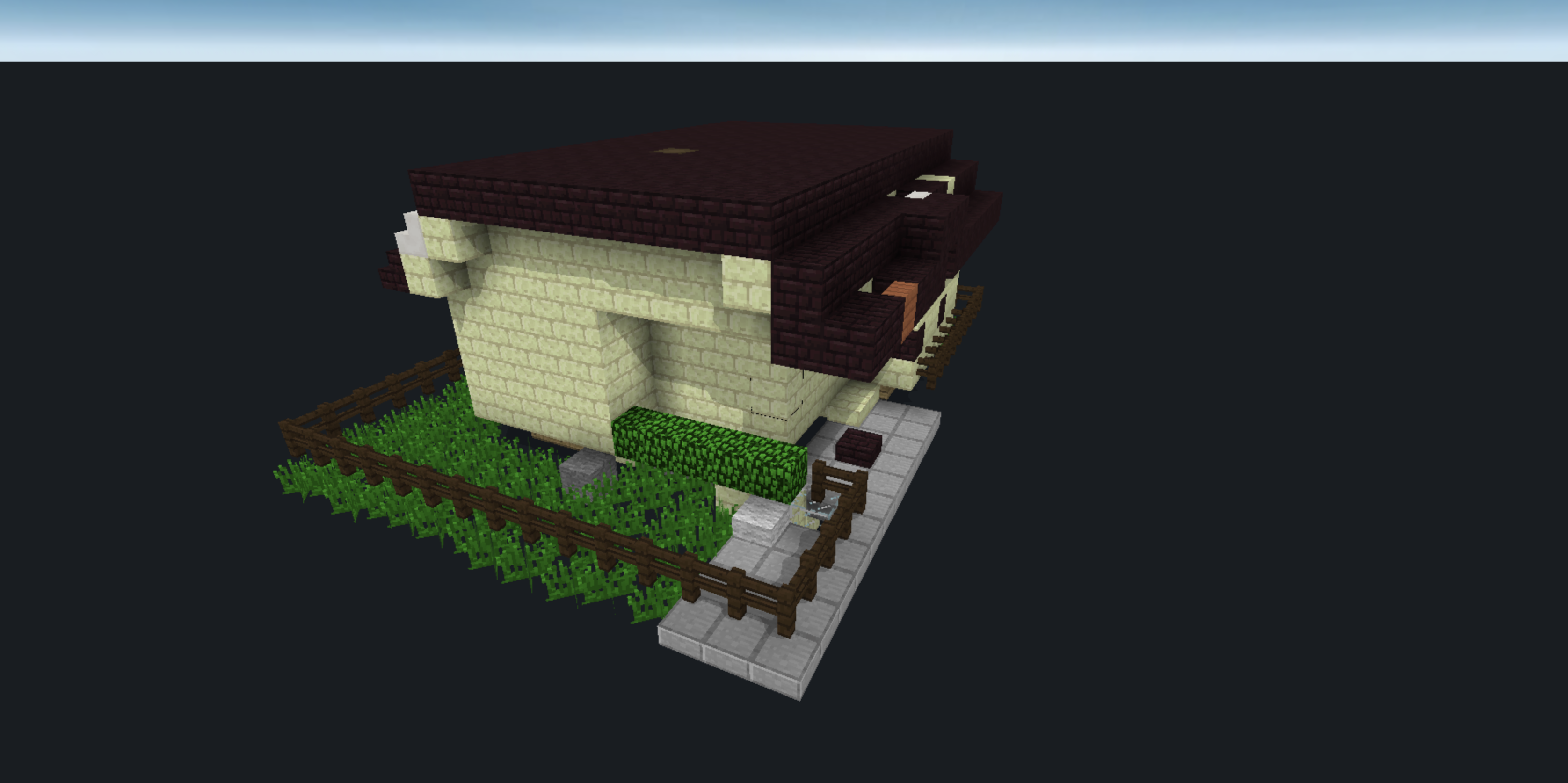} \\[2pt]
        
        \includegraphics[width=0.31\textwidth]{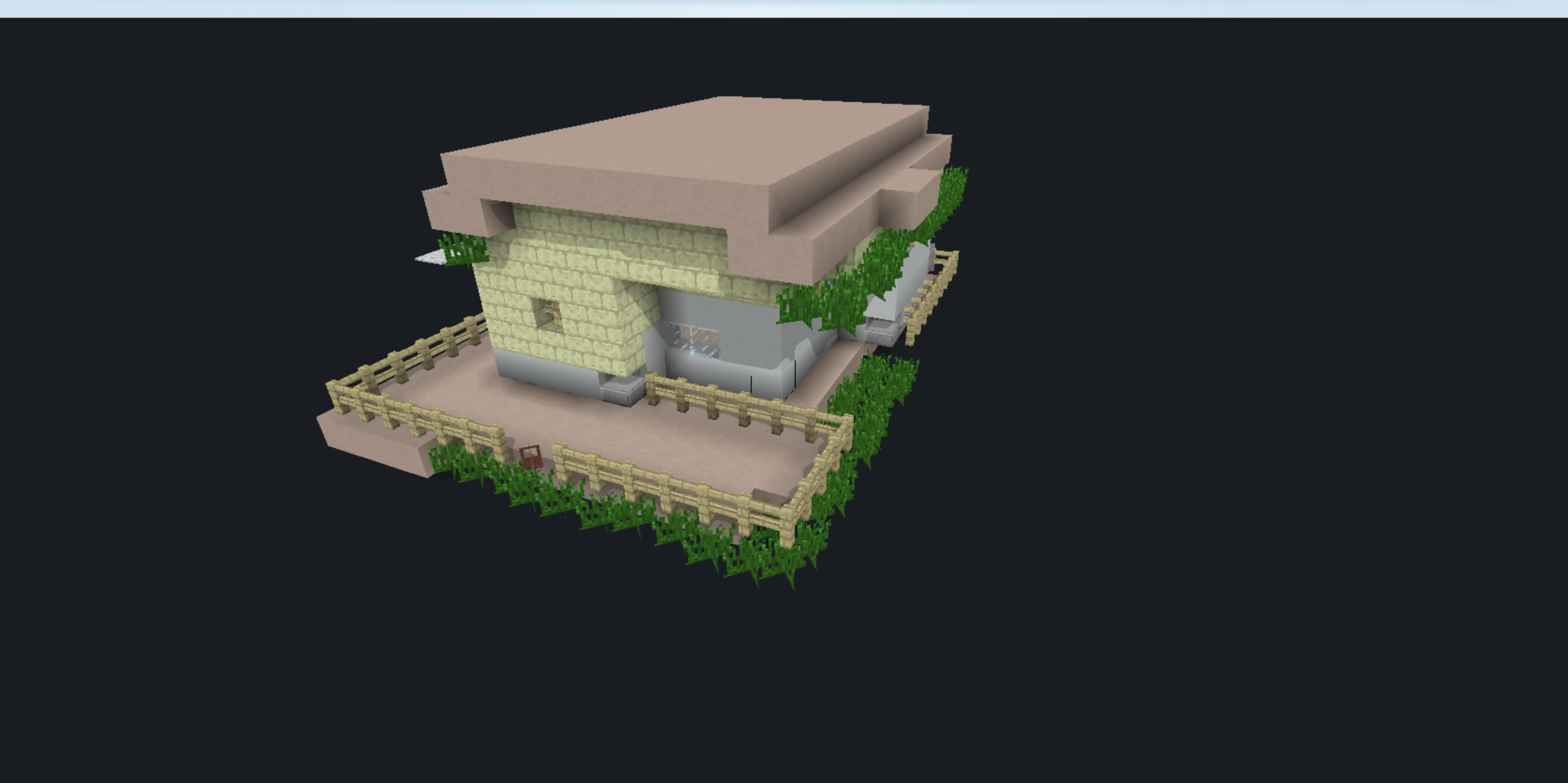} &
        \includegraphics[width=0.31\textwidth]{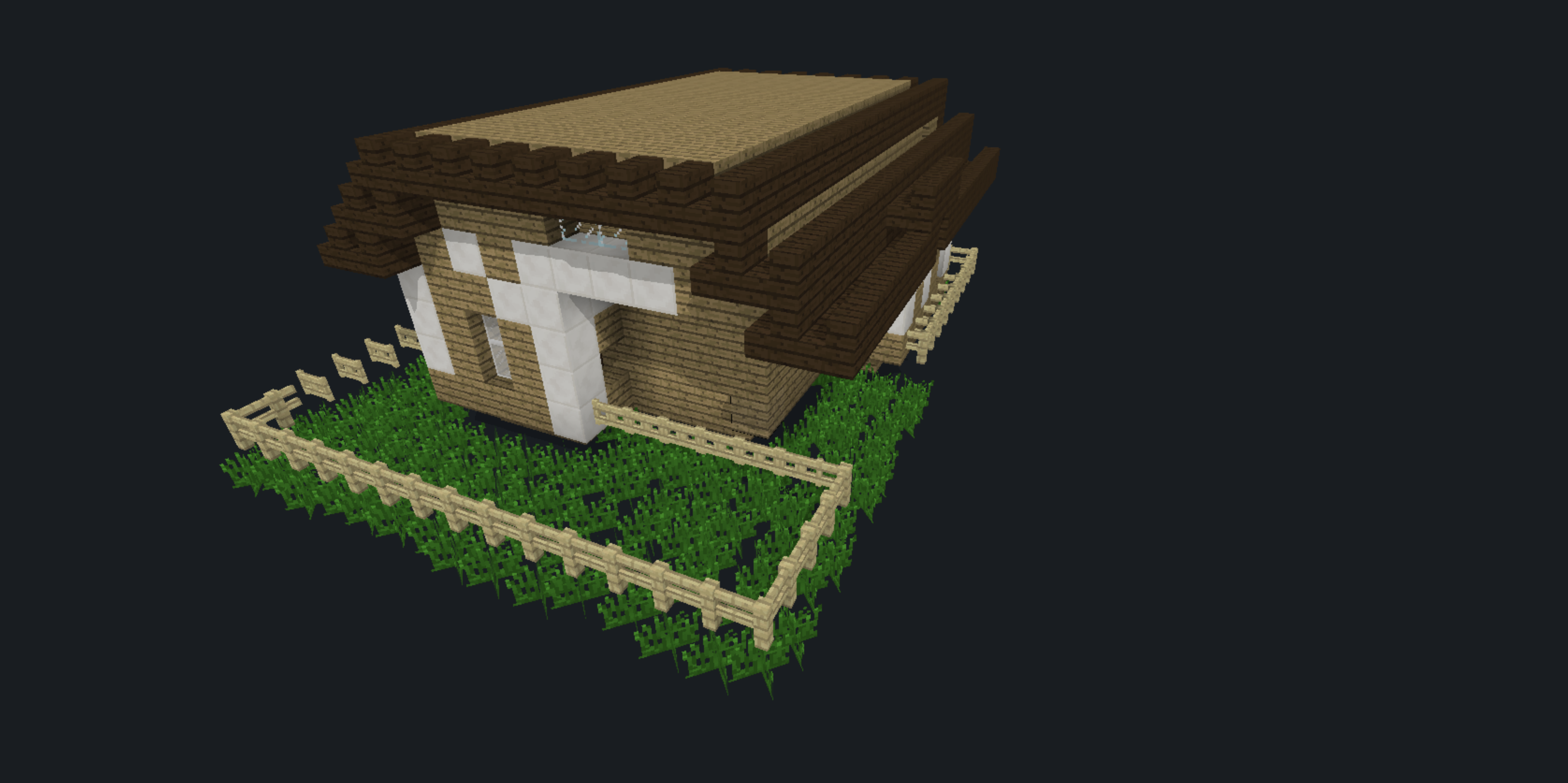} &
        \includegraphics[width=0.31\textwidth]{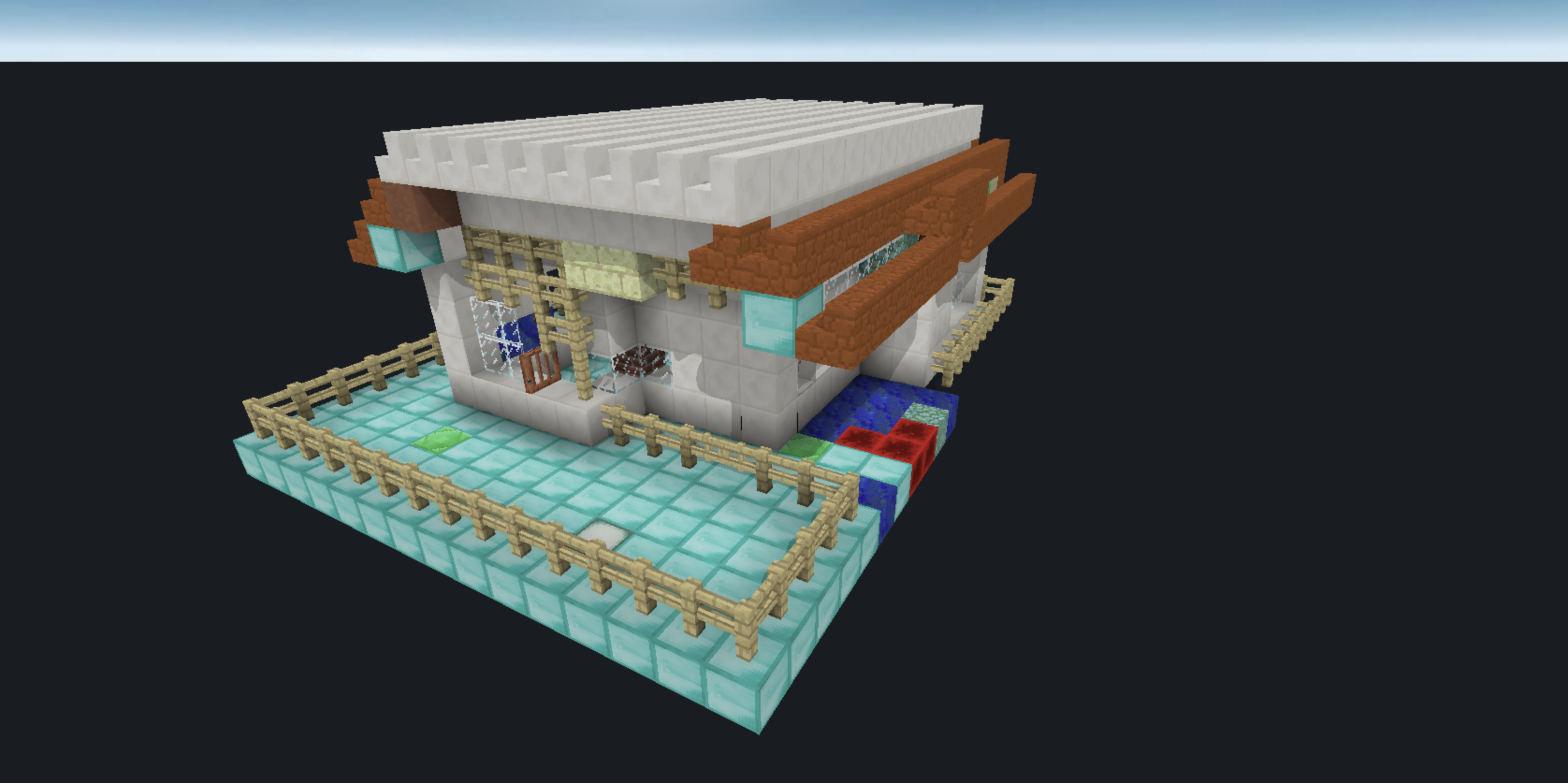} \\
    \end{tabular}
    
    \caption{Diversity of generated samples. Scaffold Diffusion produces varied and realistic 3D structures for the same occupancy map.}
    \label{fig:diverse_samples}
\end{figure*}

\subsection{Qualitative Results}
While previous works such as VoxelCNN report quantitative metrics such as perplexity and next-action accuracy in addition to qualitative results, we recognize that evaluation of generative capability is inherently a qualitative task. In Figure 2 we demonstrate the generative capability of Scaffold Diffusion in comparison to \cite{lee2023diffusion} and an autoregressive version of the model. We note that Scaffold Diffusion can generate spatially consistent and functional structures, whereas the autoregressive version typically generates collapsed structures that contain only a few different block categories types. \cite{lee2023diffusion} suffers from an overrepresentation of background tokens, which we hypothesize is due to its training on the entire voxel structure, even with inverse class-frequency loss re-weighting and VQ-VAE latent diffusion. 

Scaffold Diffusion also demonstrates diverse generative capability. In Figure 3 we illustrate the diversity of structures that Scaffold Diffusion generates for the same conditional occupancy map. 

Because qualitative evaluation can be difficult to judge based on a few samples in a figure, we create and share a live visualization demo where users can view uncurated generated samples and the diffusion generation process. The demo link is provided here: \url{https://scaffold.deepexploration.org/}. 
\subsection{Ablations}
We ablate our backbone and design decision to operate on a sequence of occupied token positions and consider operating on the entire voxel map, similar to \cite{lee2023diffusion}. We adopt DiT-3D \cite{mo2023dit} which voxelizes point clouds into voxel patches and use our voxel map directly to construct patches for the DiT transformer. We choose their default patch size of $p=4$ and use their DiT model with a depth of $n=12$ and $n=12$ heads. Similar to the behavior with \cite{lee2023diffusion}, even with inverse class-frequency loss re-weighting, we suffer from the problem of sparsity and background voxel class imbalance and fail to generate plausible boolean occupancy structures.

Incorporating 3D positional information is critical for our discrete diffusion model to generate spatially coherent structures. We consider as an ablation using learned positional embeddings with an embedding lookup table rather than a fixed 3D sinusoidal embedding. As reported in Table 1, we find significantly worse performance when relying on learned embeddings.

\begin{table}[ht]
\caption{Positional Embedding Ablation}
\label{tab:lm_results}
\vskip 0.15in
\begin{center}
\begin{small}
\begin{tabular}{lcc}
\toprule
Method & NLL $\downarrow$ & Perplexity $\downarrow$ \\
\midrule
Learned Position Encoding & 3.369 & 29.05 \\
3D Sinusoidal Positional Encoding & \textbf{0.58} & \textbf{1.787} \\
\bottomrule
\end{tabular}
\end{small}
\end{center}
\vskip -0.1in
\end{table}

\section{Limitations and Future work}
Scaffold Diffusion currently uses a boolean occupancy map to define the non-background voxel positions to be generated. Future work may include training a generative model to first generate boolean occupancy maps, leading to a fully generative two-stage process. 
Moreover, because we model each voxel as a token in our discrete diffusion sequence, the number of active voxels is limited by the sequence length of the discrete diffusion model. Interesting next directions include exploring hierarchical generation in order to generate much larger voxel structures. 

\section{Conclusion}
We introduced Scaffold Diffusion, a discrete diffusion model for generating sparse multi-category voxel structures. We show that in comparison to previous work and other formulations, Scaffold Diffusion is able to generate consistent and realistic 3D structures, extending the practical applicability of discrete diffusion beyond naturally sequential data into the 3D spatial domain.  


\bibliographystyle{plainnat}
\bibliography{references}

\end{document}